\documentclass[sigconf,nonacm]{acmart}

\setcopyright{none}
\acmConference[SynthIR '26]{SynthIR 2026: TODO Full Workshop Name}{TODO Month 2026}{TODO Location}

\usepackage{booktabs}
\usepackage{microtype}
\usepackage{graphicx}
\usepackage{xcolor}
\usepackage{enumitem}

\begin{document}

\title{Silent Failures in Multimodal Agentic Search:\\A Diagnostic Taxonomy and Cross-Judge Evaluation}

\author{Zhengxian Wu}
\affiliation{%
  \institution{Ant Group}
  \city{Hangzhou}
  \country{China}
}
\email{sean.wzx@antgroup.com}

\author{Junjie Gao}
\affiliation{%
  \institution{Ant Group}
  \city{Hangzhou}
  \country{China}
}
\email{jungao.gjj@antgroup.com}

\author{Kai Yang}
\affiliation{%
  \institution{Ant Group}
  \city{Hangzhou}
  \country{China}
}
\email{chengyue.yk@antgroup.com}

\renewcommand{\shortauthors}{Wu et al.}

\begin{abstract}
Multimodal agentic search systems increasingly rely on external tools to answer knowledge-intensive visual questions. 
However, existing evaluations mainly focus on final-answer accuracy and may miss failures in the search trajectory. 
In this work, we study such hidden reliability issues as silent failures.
We introduce a six-category taxonomy covering modality shortcuts, phantom grounding, wrong-evidence-right-answer cases, over-retrieval laundering, cross-modal contradiction, and provenance hallucination. 
Based on this taxonomy, we build a trajectory-level diagnostic pipeline that evaluates both answer correctness and evidence-grounding quality under a unified ReAct-style scaffold. 
Experiments on MMSearch-Plus trajectories across four frontier multimodal models show that surface accuracy consistently overestimates true trajectory-level correctness. 
We further use cross-judge validation, blank-image stress tests, and tool ablations to show that silent failures are capability-dependent and often shift rather than disappear.
\par
\noindent\textbf{Home-page:}
\par\noindent
\href{https://github.com/DingWu1021/silent-failures-multimodal-agentic-search}{\texttt{https://github.com/DingWu1021/}}
\par\noindent
\href{https://github.com/DingWu1021/silent-failures-multimodal-agentic-search}{\texttt{silent-failures-multimodal-agentic-search}}
\end{abstract}


\maketitle

\section{Introduction}
\label{sec:intro}
Multimodal agentic search has recently become an important paradigm for knowledge-intensive visual question answering\cite{wu2025mmsearch,Chng2025SenseNovaMARSEM}.
Unlike models that answer mainly from parametric knowledge, these systems interact with external search and browsing tools to gradually collect evidence, so that the final answer can be supported by traceable external information~\cite{mmsearch2024,mmsearchplus2026,mcsearch2025}.
This traceability is increasingly critical as fabricated and synthetic content proliferates across information retrieval ecosystems~\cite{SynthIR2026}.

Despite this progress, existing evaluations of multimodal agentic search remain largely answer-level\cite{Geng2025WebWatcherBN}. 
However, in multimodal agentic search\cite{Narayan2025DeepMMSearchR1EM}, the final answer is often produced through a sequence of intermediate decisions, including visual inspection, tool selection, evidence retrieval, and evidence integration. 
Errors may occur at any stage. 
Recent studies on agent systems also show that many failures are \emph{silent}~\cite{Cemri2025WhyDM}. 
Motivated by this gap, we ask a central question: \emph{what does it mean for a multimodal search agent to be truly correct?} 
To answer this question, we need to examine the full search trajectory, systematically characterize the forms of silent failures in multimodal search agents, measure how often they occur, and assess whether they can be automatically diagnosed.

In this work, we study multimodal agentic search from a diagnostic perspective. 
Our goal is to understand how current frontier agents behave when their full trajectories are evaluated for grounding and faithfulness. 
We first introduce a taxonomy of silent failures for multimodal agentic search. 
The taxonomy includes six recurring failure types: modality shortcut, phantom grounding, wrong-evidence-right-answer, over-retrieval laundering, cross-modal contradiction, and provenance hallucination. 
These categories cover failures before retrieval, during retrieval, and after retrieval. 
They are designed to identify trajectories whose final answers may appear reasonable, but whose reasoning processes are not sufficiently supported by the image or retrieved evidence.

Based on this taxonomy, we build a lightweight trajectory-level diagnostic pipeline. 
Each model is run under a unified ReAct-style framework with the same tool interfaces and logging format.
Then, an LLM judge guided by a structured evaluation rubric evaluates both answer correctness and the presence of each silent-failure category. 
To assess the reliability of judge-based diagnosis, we further introduce same-family and cross-family LLM cross-validators.

We apply this pipeline to 800 trajectories from a stratified MMSearch-Plus sample, covering multiple frontier multimodal models. 
Our results show that surface accuracy consistently overestimates trajectory-grounded correctness. 
We further find that silent failures do not simply disappear as models become stronger. 
Instead, their distribution shifts across models with different capabilities.
Our contributions are summarized as follows:
\begin{itemize}[leftmargin=*,topsep=2pt,itemsep=1pt]
    \item We formulate silent failures in multimodal agentic search as a trajectory-level reliability problem, showing that answer-level accuracy alone cannot distinguish faithful grounded reasoning from superficially correct but unsupported trajectories.

    \item We introduce a six-category taxonomy of silent failures and build a unified diagnostic pipeline that applies an LLM-judge rubric to evaluate both answer correctness and trajectory grounding.

    \item We assess judge reliability with same-family and cross-family validators, showing which findings are stable and which fine-grained labels remain judge-dependent.

    \item We conduct a study on MMSearch-Plus trajectories across frontier multimodal agents, revealing a gap between surface accuracy and true correctness rate, and showing that silent failures are capability-dependent and often shift rather than disappear.
\end{itemize}

\begin{figure*}[t]
\centering
\includegraphics[width=1.0\textwidth]{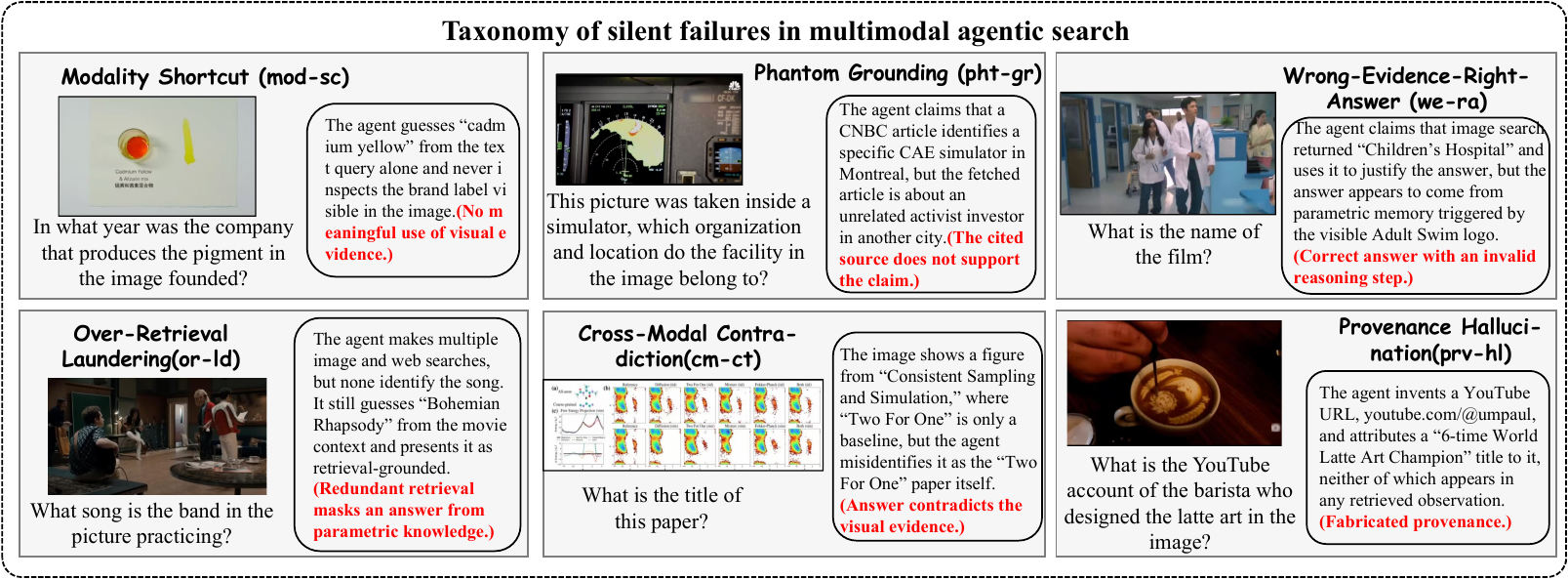}
\caption{Taxonomy of silent failures in multimodal agentic search. 
Each panel presents one failure category with a representative task example and the observed behavior that triggers the diagnostic label.}
\label{fig:main}
\end{figure*}

\section{Method}
\label{sec:method}

\subsection{Pipeline}
\label{sec:formulation}

Given a multimodal search task $x=(I,q)$, where $I$ denotes the input image and $q$ denotes the user question, an agent produces a final answer through a sequence of tool calls and intermediate reasoning steps. 
We denote one complete run as a trajectory:
\[
\tau = \{z_1, a_1, o_1, \ldots, z_T, a_T, o_T, y\},
\]
where $z_t$ is the intermediate reasoning at step $t$, $a_t$ is the action or tool call, $o_t$ is the tool observation, and $y$ is the final answer. Standard evaluation usually checks only whether $y$ matches the reference answer, while ignoring the intermediate reasoning and evidence use in $\tau$.

We argue that answer correctness alone is not sufficient to measure reliability in multimodal agentic search. 

To this end, for each trajectory, we produce a structured diagnostic record:
\[
d(\tau) = \{c, f_1, f_2, \ldots, f_6\},
\]
where $c \in \{0,1\}$ indicates whether the final answer is correct, and each $f_i \in \{0,1\}$ indicates whether the trajectory triggers the $i$-th silent-failure category. 
The failure labels are not mutually exclusive and a single trajectory may contain multiple failure modes.
This formulation extends answer-level evaluation into trajectory-level diagnosis.

\vspace{-0.3cm}
\subsection{Taxonomy of Silent Failures}
\label{sec:taxonomy}

To characterize failures in multimodal agentic search, we introduce a taxonomy of silent failures tailored to this setting.
The taxonomy contains six failure types and covers three stages of the search process: before retrieval, during retrieval, and after retrieval. 

\textbf{(1)Modality Shortcut.}
A modality shortcut occurs when the agent produces an answer without sufficiently using the input image. 
The model may make no image-related tool calls, or it may only describe the image at a superficial level before relying mainly on the text question or its own parametric knowledge. 
This failure usually occurs before retrieval. 
It indicates that the model bypasses the multimodal nature of the task instead of using the visual information.
\textbf{(2)Phantom Grounding.}
Phantom grounding occurs when the agent cites a retrieved source, but the source content does not actually support the claimed fact. 
This failure is more subtle than the absence of retrieval, because the trajectory does contain external evidence. 
However, there is no valid support relation between the evidence and the conclusion. 
For example, the model may cite a webpage as evidence even though the page only contains information unrelated to the answer.
\textbf{(3)Wrong-Evidence-Right-Answer.}
This failure refers to cases where the final answer is correct, but some explicit evidence or reasoning step in the trajectory is factually wrong.
This failure shows that a correct answer does not necessarily imply a reliable reasoning process. 
The model may arrive at the correct answer through parametric knowledge or accidental cues, while still providing an incorrect retrieval-based explanation or an invalid reasoning chain.
\textbf{(4)Over-Retrieval Laundering.}
Over-retrieval laundering occurs when the agent issues multiple redundant retrieval calls, but the final answer is not supported by the retrieved results and instead comes from the model's prior parametric knowledge. 
In this case, retrieval mainly serves as a form of presentation, making the answer appear externally grounded. 
This failure often appears during the retrieval process, especially when the model gives a detailed and confident answer after unhelpful retrieval results.
\textbf{(5)Cross-Modal Contradiction.}
Cross-modal contradiction occurs when the final answer agrees with some noisy text retrieval results but conflicts with what is clearly shown in the input image. 
This failure reflects a central challenge in multimodal search: the model must not only retrieve evidence, but also check consistency between visual and textual evidence. 
When the model over-relies on noisy text evidence and underuses the image, this type of failure can arise.
\textbf{(6)Provenance Hallucination.}
Provenance hallucination occurs when the agent generates URLs, dates, source names, or citations that look plausible but do not appear in any tool observation. 
This failure directly harms the traceability of the answer. Even if the final answer appears credible, invented provenance can mislead users into believing that the answer is supported by real external evidence.

These six categories are not mutually exclusive.
Instead, they capture different aspects of trajectory reliability.
Modality shortcut mainly corresponds to pre-retrieval failure, over-retrieval laundering mainly corresponds to retrieval-process failure, and phantom grounding, wrong-evidence-right-answer, cross-modal contradiction, and provenance hallucination mainly correspond to post-retrieval failures. 
With this taxonomy, we can identify hidden reliability issues that are difficult to detect from the final answer alone.
\vspace{-0.3cm}
\subsection{Trajectory-Level Diagnostic Pipeline}
\label{sec:pipeline}

Based on this taxonomy, we build a lightweight diagnostic pipeline. 
The pipeline consists of three steps: unified trajectory collection, rubric-guided LLM diagnosis, and cross-judge validation.

\textbf{Unified trajectory collection.}
To ensure fair comparison across models, we run all models under the same ReAct-style agent framework. 
Each model uses the same system prompt, the same tool interfaces, and the same trajectory logging format. The tool set includes web search, image search, webpage reading, and image cropping. 
For each task--model pair, we record the full reasoning, tool calls, tool inputs, tool observations, and final answer. 
Each trajectory is allowed a fixed maximum number of tool calls, which prevents unbounded search and keeps the budget consistent across models.

\textbf{Rubric-guided LLM diagnosis.}
After collecting the trajectory, we use an LLM judge guided by a structured evaluation rubric to diagnose it. 
The judge receives the task question, the input image, the reference answer, the full trajectory, and the final answer. 
It then produces a structured output that includes whether the final answer is correct and whether each of the six silent-failure categories is present.
For each triggered failure category, the judge also provides a short justification and points to the relevant trajectory steps. This improves the interpretability of the diagnosis and supports later manual inspection and agreement analysis.
Based on the diagnostic output, we define the true correctness rate as requiring both a correct final answer and no silent-failure flags in the trajectory.

\textbf{Cross-judge validation.}
Since LLM judges may have their own biases, we introduce a cross-validation mechanism to assess the stability of the diagnostic results. 
Specifically, we use both a same-family judge and a cross-family judge to re-evaluate the same stratified set of trajectories, and compute Cohen's $\kappa$ between each validator and the primary judge.

This design serves two purposes. 
First, it tests whether answer-correctness judgments remain stable across different judges. 
Second, it measures whether fine-grained failure categories depend on the choice of judge. 
\begin{table}[tb]
\centering
\footnotesize
\setlength{\tabcolsep}{1pt}
\caption{Trajectory terminal states per model ($N=200$ each).
Headline metrics are computed on the committed subset.}
\label{tab:states}
\begin{tabular}{lrrrr}
\toprule
\textbf{Model} & \textbf{Committed} & \textbf{Refused} & \textbf{Exhausted} & \textbf{Crashed} \\
\midrule
Claude Sonnet 4.6        & 136 (68\%) & 0 (0\%) & 46 (23\%) & 18 (9\%) \\
Gemini 2.5 Pro           & 133 (66\%) & 1 (0\%) & 58 (29\%) &  8 (4\%) \\
Gemini 3.1 Pro Preview   & 124 (62\%) & 0 (0\%) & 76 (38\%) &  0 (0\%) \\
GPT-4o                   & 175 (88\%) & 3 (2\%) & 14 (7\%)  &  8 (4\%) \\
\bottomrule
\end{tabular}
\end{table}
\section{Experiments}
\label{sec:experiments}

\subsection{Experimental Setup}
\label{sec:exp-setup}

\textbf{Data and models.}
We sample 200 tasks from MMSearch-Plus using stratified sampling, and run a unified search-agent framework on four frontier multimodal models, resulting in 800 trajectories in total. The sampling covers different task categories and difficulty levels, so that the results are not dominated by a single task type. The evaluated models include Claude Sonnet~4.6, Gemini~2.5~Pro, Gemini~3.1~Pro~Preview, and GPT-4o.
\textbf{Trajectory collection.}
All models are run under the same ReAct-style scaffold, with identical tool interfaces, system prompts, and trajectory logging format. Each trajectory is allowed at most 10 tool calls. We record the full intermediate reasoning, tool calls, tool observations, and final answer. 
\textbf{Trajectory states.}
Not every trajectory produces a substantive answer that can be evaluated. We therefore divide terminal states into four types:
\emph{committed}, where the model gives a substantive final answer;
\emph{refused}, where the model refuses to answer;
\emph{exhausted}, where the model reaches the tool-call limit without producing a final answer;
and \emph{crashed}, where the model run fails due to an unrecoverable error.
Since our goal is to study reliability when the model actually attempts to answer, we report the main results on the committed subset, while also reporting raw-$N$ results as additional context as shown in Table~\ref{tab:states}.
\textbf{Metrics.}
We report two main metrics.
\emph{Accuracy} measures whether the final answer is correct.
\emph{True Correctness Rate} (TCR) further requires the trajectory to be answer-correct and free of any silent-failure flag.
Thus, Accuracy measures surface answer correctness, while TCR measures trajectory-level correctness. 

\subsection{Main Results}
\label{sec:main-results}
Table~\ref{tab:acc-vs-tcr} reports Accuracy and TCR for the four models on the committed subset. Across all models, TCR is lower than surface Accuracy. After removing answer-correct trajectories with silent-failure flags, the measured correctness drops by 0.6 to 7.3 percentage points: Claude drops from 36.0\% to 34.6\%, Gemini~2.5 from 31.6\% to 28.6\%, Gemini~3.1 from 61.3\% to 54.0\%, and GPT-4o from 22.9\% to 22.3\%.

These results show that answer-level accuracy overestimates the reliability of multimodal search agents. Even when the final answer is judged correct, the full trajectory may still contain problems that answer-level evaluation cannot capture, such as wrong evidence, hallucinated provenance, or cross-modal contradiction. Notably, Gemini~3.1 obtains the highest committed accuracy, but also has the largest drop from Accuracy to TCR. This suggests that stronger models do not necessarily eliminate silent failures. As models produce correct answers more often, cases where the answer is correct but the process is unreliable can also become more visible.


\begin{table}[tb]
\centering
\footnotesize
\setlength{\tabcolsep}{3pt}
\caption{Accuracy versus true correctness rate (TCR) on the committed subset, with 95\% Wilson CIs.}
\label{tab:acc-vs-tcr}
\begin{tabular}{lrrrr}
\toprule
\textbf{Model} & \textbf{Raw $N$} & \textbf{Acc (raw)} & \textbf{Acc (committed)} & \textbf{TCR (committed)} \\
\midrule
Claude Sonnet 4.6      & 200 & 25.0\% & 36.0\% [28.4, 44.4] & 34.6\% [27.1, 42.9] \\
Gemini 2.5 Pro         & 200 & 21.5\% & 31.6\% [24.3, 39.9] & 28.6\% [21.6, 36.8] \\
Gemini 3.1 Pro Preview & 200 & \textbf{39.0\%} & \textbf{61.3\% [52.5, 69.4]} & \textbf{54.0\% [45.3, 62.6]} \\
GPT-4o                 & 200 & 20.0\% & 22.9\% [17.3, 29.6] & 22.3\% [16.8, 29.0] \\
\bottomrule
\end{tabular}
\end{table}

\subsection{Diagnostic Analyses}
\label{sec:diagnostic-analyses}

\begin{figure}[tb]
\centering
\includegraphics[width=\linewidth]{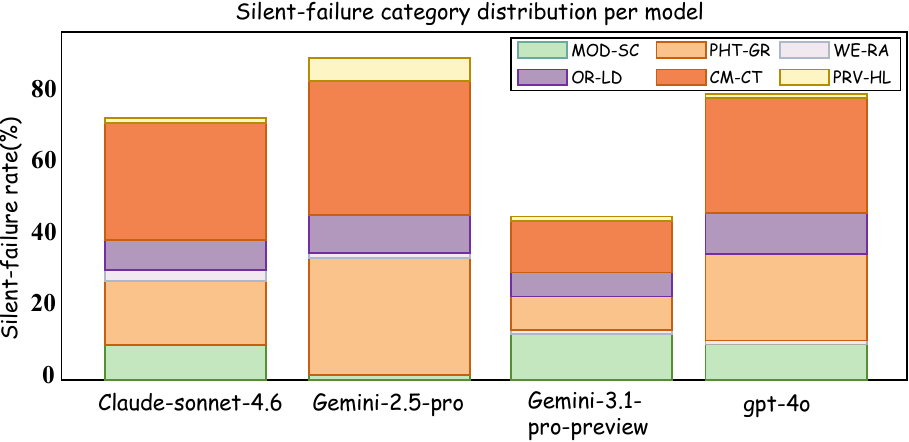}
\caption{Silent-failure rates by category on the committed subset, revealing capability-dependent failure shifts.}
\label{fig:failure-distribution}
\vspace{-0.3cm}
\end{figure}

\textbf{Failure profiles shift with model capability.}
To understand how different models fail silently, we analyze the distribution of failure categories among committed trajectories, as shown in Figure~\ref{fig:failure-distribution}. 
The results show that silent failures do not simply decrease as model capability improves; instead, the failure profile shifts. 
For Claude, Gemini~2.5, and GPT-4o, failures mainly occur after retrieval, especially \textsc{pht-gr} and \textsc{cm-ct}. 
This suggests that these models often perform retrieval, but still fail to use the retrieved evidence reliably. 
For example, they may cite webpages that do not support the final answer, or follow textual evidence that conflicts with the image.
Gemini~3.1 shows a different pattern.
It has the lowest rates of \textsc{pht-gr} and \textsc{cm-ct}, indicating fewer post-retrieval evidence-use failures, but it has the highest rate of \textsc{mod-sc}.
Our trajectory inspection shows that Gemini~3.1 identifies the image with its vision encoder and answers without calling reverse image search. 
When visual recognition is incorrect, this behavior becomes a pre-retrieval modality shortcut. 
Thus, stronger models do not eliminate silent failures; instead, they shift where failures occur.

\textbf{Blank-image stress test.}
One possible concern is that some MMSearch-Plus questions may not truly require the image. If correct answers can be obtained without the image, then the trajectory issues we observe may not necessarily reflect failures in multimodal grounding. To test this, we perform a blank-image stress test on all trajectories in the committed subset that are judged answer-correct by the primary judge.
Specifically, we replace the input image with a same-size blank white image, rerun the model under the same scaffold, and judge the resulting trajectory again. This test covers 208 originally answer-correct model--task pairs, including 50 for Claude, 43 for Gemini~2.5, 75 for Gemini~3.1, and 40 for GPT-4o.
As shown in Table~\ref{tab:blank-image}, the survival rate under blank images is zero for three models, and only 2/75 for Gemini~3.1. Overall, 206 out of 208 correct trajectories no longer remain after removing the image.
We further inspect the two surviving cases and find that they are not cases where the model successfully bypasses the image. Instead, the question text itself already contains enough information. 

\begin{table}[tb]
\centering
\caption{Blank-image stress test on originally correct committed trajectories, reported as survival cases.}
\label{tab:blank-image}
\small
\begin{tabular}{lcccc}
\toprule
Model & Claude & Gemini 2.5 & Gemini 3.1 & GPT-4o \\
\midrule
Survival rate & 0/50 (0.0\%) & 0/43 (0.0\%) & 2/75 (2.7\%) & 0/40 (0.0\%) \\
\bottomrule
\end{tabular}
\vspace{-0.4cm}
\end{table}


\textbf{Judge reliability.}
Table~\ref{tab:judge-agreement} reports cross-judge agreement on the same set of $n=100$ paired verdicts. Answer correctness is highly stable across judges, with $\kappa=0.916$ for the same-family validator and $\kappa=0.817$ for the cross-family validator. The cross-family observed agreement $P_o$ is also high at 98\%. 

For fine-grained failure categories, the pattern is more mixed. \textsc{mod-sc} has the lowest cross-family agreement, with $\kappa=0.031$ and $P_o=25\%$, indicating a strong difference in how judges apply this label. \textsc{or-ld} shows a similar but weaker pattern, with cross-family $\kappa=0.301$ and $P_o=66\%$. These results suggest that some categories are sensitive to the judge's implicit decision boundary, especially when deciding whether the model truly used the image or whether retrieval mainly served to support parametric knowledge.

Other categories show different behavior. \textsc{pht-gr} and \textsc{cm-ct} have moderate cross-family observed agreement, with $P_o=75\%$ and $79\%$, respectively, suggesting more balanced judgment differences. In contrast, \textsc{prv-hl} and \textsc{we-ra} have very high $P_o$ but few positive cases, so their $\kappa$ are unstable or undefined under low prevalence. 

\begin{table}[tb]
\centering
\footnotesize
\setlength{\tabcolsep}{10pt}
\caption{Cross-judge agreement on $n=100$ paired verdicts, reporting Cohen's $\kappa$ for same- and cross-family validators and cross-family observed agreement $P_o$.}
\label{tab:judge-agreement}
\vspace{-0.3cm}
\begin{tabular}{lccc}
\toprule
\textbf{Metric} 
& \textbf{Same $\kappa$} 
& \textbf{Cross $\kappa$} 
& \textbf{Cross $P_o$} \\
\midrule
answer\_correct & 0.916 & 0.817 & 98\% \\
\textsc{mod-sc} & 0.260 & 0.031 & 25\% \\
\textsc{pht-gr} & 0.620 & 0.260 & 75\% \\
\textsc{or-ld}  & 0.565 & 0.301 & 66\% \\
\textsc{cm-ct}  & 0.513 & 0.467 & 79\% \\
\textsc{prv-hl} & 0.322 & 0.322 & 96\% \\
\textsc{we-ra}  & --    & --    & 99\% \\
\bottomrule
\end{tabular}
\vspace{-0.2cm}
\end{table}

\textbf{Tool ablation: better tools reshape rather than eliminate failures.}
Finally, we study how tool design affects failure patterns by comparing a full setting with \texttt{reverse\_image\_search} against an ablated setting where this tool is removed, while keeping all other tools, prompts, and execution settings unchanged. 
As shown in Table~\ref{tab:v1-v2}, \texttt{reverse\_image\_search} improves committed accuracy by 13.9 to 18.9 percentage points for the three more retrieval-dependent models: Claude, Gemini~2.5, and GPT-4o. 
It also shifts their failure distribution: \textsc{mod-sc} and \textsc{or-ld} decrease, suggesting less image bypassing and less use of redundant retrieval to support parametric knowledge, while \textsc{pht-gr} and \textsc{cm-ct} increase, suggesting that models obtain more image-related evidence but may still misuse it.
Gemini~3.1 shows a different pattern: the tool improves committed accuracy by only 4.8 points and only slightly reduces all silent-failure categories. 
Trajectory inspection suggests that Gemini~3.1 often identifies the image with its own vision encoder before any tool call, so reverse image search largely duplicates an existing capability while consuming tool-call budget. 
Overall, better tools do not always reduce failures monotonically; their effect depends on model capability and often changes where failures occur.

\begin{table}[tb]
\centering
\small
\caption{Controlled tool ablation with versus without \texttt{reverse\_image\_search} on the committed subset.}
\label{tab:v1-v2}
\vspace{-0.3cm}
\begin{tabular}{lrrrr}
\toprule
& \textbf{Claude} & \textbf{Gem 2.5} & \textbf{Gem 3.1} & \textbf{GPT-4o} \\
\midrule
Acc.\ v1        & 19.6 & 12.7 & 56.5 & 8.6 \\
Acc.\ v2        & 36.0 & 31.6 & 61.3 & 22.5 \\
$\Delta$Acc.    & $+16.5$ & $+18.9$ & $\mathbf{+4.8}$ & $+13.9$ \\
\midrule
$\Delta$\textsc{mod-sc} & $-10.1$ & $-6.5$ & $\mathbf{-0.3}$ & $-23.2$ \\
$\Delta$\textsc{or-ld}  & $-35.4$ & $-15.6$ & $\mathbf{-1.6}$ & $-1.9$ \\
$\Delta$\textsc{pht-gr} & $+1.8$  & $+16.6$ & $\mathbf{-2.0}$ & $+18.9$ \\
$\Delta$\textsc{cm-ct}  & $+9.1$  & $+8.3$  & $\mathbf{-2.0}$ & $+13.5$ \\
\bottomrule
\end{tabular}
\vspace{-0.4cm}
\end{table}

\vspace{-0.2cm}
\section{Conclusion}
\label{sec:conclusion}
We presented a trajectory-level study of silent failures in multimodal agentic search. Our taxonomy and diagnostic pipeline show that final-answer accuracy is insufficient for measuring reliability, since correct answers may still rely on unsupported evidence, hallucinated provenance, or cross-modal contradictions. Experiments on MMSearch-Plus reveal a consistent gap between surface accuracy and true correctness rate. Cross-judge validation shows that answer correctness is stable, while fine-grained failure labels are more judge-dependent. Blank-image and tool-ablation analyses further suggest that stronger models and better tools often shift failures rather than remove them.

\bibliographystyle{ACM-Reference-Format}
\bibliography{refs}

\end{document}